\documentclass[11pt]{article}
\usepackage{paclic32}
\usepackage{times}
\usepackage{latexsym}
\usepackage{amsmath}
\usepackage{multirow}
\usepackage{url}
\usepackage{color}

\title{CVIT-MT Systems for WAT-2018}
\author{Jerin Philip$^\dagger$,  Vinay P. Namboodiri$^\ddagger$ and C.V. Jawahar$^\dagger$ \\
  $^\dagger$ CVIT, IIIT Hyderabad, $^\ddagger$ IIT-Kanpur \\
  {\tt jerin.philip@research.iiit.ac.in} \\
  {\tt vinay.pn@cse.iitk.ac.in, jawahar@iiit.ac.in }
}
\date{}

\begin{document}
\maketitle
\begin{abstract}
  This document describes the machine translation system used in the submissions of IIIT-Hyderabad ({\sc cvit-mt}) for the \textsc{wat-2018} English-Hindi translation task. Performance is evaluated on the associated corpus provided by the
  organizers. We experimented with convolutional sequence to sequence architectures. We also train with additional data obtained through backtranslation.
\end{abstract}

\section{Introduction}

Innovations in Neural Machine Translation (\textsc{nmt}) have led to success in many machine translation tasks, often outperforming Statistical Machine Translation ({\sc smt}) techniques. Similar to many other language pairs, {\sc nmt} based approaches have been attempted 
for the English-Hindi language pair as
well (e.g. the {\sc wat-2017} submission ~\cite{wang2017xmu}). Hindi  continue to remain as a low resource language demanding further attention from Natural Language Processing ({\sc nlp}, Machine Learning {\sc ml} and other related communities. The Hindi-English pair has limited availability of sentence level aligned bitext as parallel corpora.

% While it is possible to give an exact translation of Hindi to English in most cases, the problem is ill-posed in the other direction - possibly leading to ambiguity in predicting Hindi from English. The only way to resolve such ambiguity would be a larger context.

Lack of sufficient data for Indian languages motivated us to explore techniques that can help in low-resource situations.
Recent works (such as \cite{edunov2018understanding,lample2017unsupervised}) point to the use of iterative backtranslation to improve translation in low resource languages or under the unavailability of parallel corpora. 

This paper describes an overview of the submission from  IIIT Hyderabad ({\sc cvit-mt}) in {\sc wat-2018}\cite{nakazawa-EtAl:2018:WAT2018} for for the Hindi-English and English-Hindi translation tasks of the mixed domain tasks. In Section~\ref{section:system-description}, we describe the components constituting our pipeline, following which in
Section~\ref{section:experimental-setup} we provide the details of the data used and procedure used for the training. Section~\ref{section:summary} summarizes our results for {\sc wat-2018}. Finally in Section ~\ref{section:additional-transformer-experiments} we include additional results using newer architectures. We conclude our observations in Section ~\ref{section:observations}.

\section{System Description}
\label{section:system-description}

In this section, we describe the details associated with the tokenization, architecture and data augmentation. These are the three components that helped in obtaining superior results on the corpus provided by the organizers of {\sc wat-2018}.
%via backtranslation used in the pipeline.

\subsection{Tokenization}

A popular method of addressing rare-words without compromising coverage of the entire corpus was Byte Pair Encoding (\textsc{bpe}) \cite{sennrich2016neural}, which used a deterministic greedy compression based algorithm to bring the vocabulary down to a finite feasible value. 

% The notion of subwords which are either frequent full words or comprised of sub-units of words  which are statistically significant helped improve translations by avoiding out of vocabulary words. 

SentencePiece \cite{kudo2018spiece}  builds on top of byte pair encoding. Unlike \textsc{bpe}, which is agnostic to language, SentencePiece gives the most likely derivation of a sentence composed of subword units. This setting reduces to character level in case a completely unknown sentence/word is provided, and the  translation model also learns to transliterate. We use SentencePiece for its merits mentioned above.

\subsection{Convolutional Sequence to Sequence Learning }

In our submission, we employ the Convolutional Sequence to Sequence architecture (\textsc{convs2s}) \cite{gehring2017convolutional}. \textsc{convs2s} follows an encoder decoder architecture. This has the advantage of being faster than the popular Recurrent Neural Network (\textsc{rnn}) based encoder decoder architectures with attention. This is because the context is built through multiple inputs stacking $k$ convolution blocks ($O(\frac{n}{k})$) with the ability to build in parallel representations for multiple parts of the sentence, unlike through time in the \textsc{rnn} ($O(n)$).

A 1-D convolutional filter of width $w$ with two channels at the output sliding over the embeddings of the text inputs constitute a basic convolutional block. Output of one channel builds up context representation and the other is used to enable gating through Gated Linear Units (\textsc{glu}s) \cite{dauphin2017language}. The encoder is constructed by stacking $k$ of the above setup, creating a receptive field controlled by $w$ and $k$. The decoder is similar to the encoder in architecture, with a fully connected layer projecting output to vocabulary size.

%\subsection{Beam Search}

\subsection{Backtranslation}

Backtranslation is a widely tried and tested data augmentation method, proposed for aiding {\sc nmt} in languages low on parallel resources using available monolingual data 
by Sennrich {\em et. al} \shortcite{sennrich2016improving}. The method works by first training a model in the low to high resource direction followed by using this model on monolingual data. The process provides more authentic sentences in the resource-scarce language and close approximation of its translation in the high resource language. It has been empirically shown that synthetic data alone generated through backtranslation can attain upto 83\% of the performance using proper bitext \cite{edunov2018understanding}.

In the next section, we describe how the components explained above are implemented and used in training - including generating dataset, preprocessing and filtering the training samples, hyperparameters of the architectures in place and evaluations.
\section{Experimental Setup}
\label{section:experimental-setup}
\subsection{Dataset}

In our experiments, we use the training data provided by organizers. In addition, we also use data obtained from translated Hindi content available on Internet.
%English through google translate. 
Top level statistics of the data used are provided in Table~\ref{table:dataset}.

\begin{table}[htbp]
    \centering
    \begin{tabular}{|l|r|c|c|}
    \hline
         \textbf{Dataset} & \textbf{Pairs} & \multicolumn{2}{c|}{\textbf{Tokens}} \\ \cline{3-4}
         & & hi & en \\ \hline
        IITB train & 1,492,827 & 22.2M & 20.6M \\
        IITB train$^\dagger$ & 923,377 & 20.3M & 18.9M  \\
        National News & 2,495,129 & 41.2M & 39.0M \\ 
        Backtranslated &  5,653,644 & 77.5M & 91.9M  \\\hline
        IITB dev & 505 & 10,656& 10,174 \\
        IITB test & 2,507& 49,394 &57,037 \\
        \hline
    \end{tabular}
    \caption{Descriptions of the corpora used, IITB train$^\dagger$ is a filtered version of the IITB train corpus.}
    \label{table:dataset}
\end{table}

The training corpus provided by the organizers, hereafter denoted by IITB-corpus consists of data from mixed domains. There are roughly 1.5M samples in training data from diverse sources, while the development and test sets are from newspaper crawls. In addition to this, monolingual data collected by the organizers from several sources are used in our backtranslation enabled attempts at training an \textsc{nmt} system. There are ~45M samples in the monolingual corpus provided.

We enhanced the training data with 
additional pairs, but automatically
translated. Note that no manual translation was used to create additional data.
We obtain 2.5M Hindi sentences automatically translated to English 
%by Google's online available translation module, 
from newspapers and similar resources,
obtained from Internet. This data is some what domain specific. They
are primarily, from news articles related to national news. This is mentioned as National News in 
Table  ~\ref{table:dataset}.

We also create a parallel corpus through backtranslation using the organizers monolingual Hindi data hereafter denoted by Backtranslated, the details of which are also included in Table~\ref{table:dataset} and the methods of creation elaborated in Section \ref{subsection:training}.

\subsection{Data Processing}

     We train separate SentencePiece models using official implementation available online \footnote{\url{https://github.com/google/sentencepiece}} with vocabulary restricted to 8000 units to function as a learned tokenizer for both English and Hindi. We use the unigram model, which gives language aware tokenization.

To filter any noisy content from IITB corpus, \emph{langdetect}\footnote{\url{https://github.com/Mimino666/langdetect}} and removed every pair which had probability of being in the respective language less than 0.95. This gave us roughly 0.92M pairs for training, from IITB corpus and is indicated as IITB train$^\dagger$ in Table \ref{table:dataset}. English data is kept true-cased,  which we found to have better results consistently with our {\sc nmt} model.

\begin{table*}
\centering
\begin{tabular}{|l|c|c|c|c|c|c|c|c|}
    \hline 
    Dataset & \multicolumn{4}{c|}{en-hi} & \multicolumn{4}{c|}{hi-en} \\ \cline{2-9}
     & BLEU & RIBES & AM-FM & Human & BLEU & RIBES & AM-FM & Human  \\ \hline
    IITB train$^\dagger$ & 13.25 & 0.695113 & 0.647220 & - & 11.83 & 0.675462 & 0.572900 & -\\
    National News & 18.77 & 0.748008 & 0.697630 & -  & 19.53	 &0.745764 & 0.614260 & - \\ 
    ~+IITB train$^\dagger$ & 19.69 & 0.758365	& 0.699810  &69.50 &  20.63 & 0.751883 & 0.623240 &  72.25	 \\ 
    Backtranslated & 16.77	& 0.714197	& 0.664330 &50.50	 & - & - & - & -\\\hline\hline
       % national + IITB & 21.10 & 0.771549	& 0.712200  & - &  20.63 & \textbf{0.751883} & 0.623240 &  \textbf{72.25}	 \\ 
    2017 Best  & 21.39 & 0.749660 & 0.688770 & 64.50 & 22.44 & 0.750921 & 0.629530 & 68.25 \\
    2018 Best & 20.28 & 0.761582 & 0.704220 & 77.00 & 17.80	 & 0.731727 & 0.611090 & 67.25\\
    \hline
\end{tabular}
\caption{Quantitative results of translating English to Hindi and vice versa.}
\label{table:results}
\end{table*}

\subsection{Training}
\label{subsection:training}

In our experiments we use the fairseq \footnote{\url{http://github.com/pytorch/fairseq} (formerly fairseq-py) } toolkit. For the tasks in this submission we use the \textsc{convs2s} model.

The encoder and decoder embeddings have a dimension of 512. The hidden units in the encoder and decoder are also 512 dimensional, following Gehring {\em et. al} \shortcite{gehring2017convolutional}. We use convolutional filters of width 3 and 20 layers stacked for both the encoder and decoder. A dropout with probability 0.1 is put in-place right after the embeddings layer for better generalization. The training is run in batches of maximum 4000 tokens at a time, which is on an average 140 sample sentences per batch. The model is trained to minimize the categorical cross-entropy loss at the token level using Nestorov accelerated gradient descent. Decoding is performed through beam search with a beam width of 10. 

% A delayed gradient update every 16 steps gives an effective batch size of 2000 on a smaller GPU setup.

We run training using four NVIDIA 1080Ti-s until validation loss hasn't improved for 3 epochs straight. The training time was roughly 2 days and stopping around 30-40 epochs.

We keep our model hyperparameters constant as specified across experiments and work with different combinations of corpora created from augmenting the National News dataset and official parallel corpora. For creating the Backtranslated corpus, we use a model trained to translate from Hindi to English using both National News and IITB corpus. We filter the obtained pairs using confidence of translation obtained from the beam-score and further to pairs with a length between 10 and 30 tokens.

\subsection{Evaluations}

We report Bilingual Evaluation Understudy (BLEU) \cite{papineni2002bleu}, Rank-based Intuitive Bilingual Evaluation Score (RIBES) \cite{isozaki2010automatic}, Adequacy-fluency metrics (AM-FM) \cite{banchs2015adequacy} for all our attempts and scores from WAT-2018 human evaluations(Human in Table \ref{table:results}) when available.

BLEU is computed as the geometric mean of unigram, bigram, trigram and 4-gram precision multiplied by a brevity penalty (BP). BLEU ranges from 0 to 1, but the values reported in Tables \ref{table:results} and \ref{table:transformer-results} are in percentages. RIBES, also giving a value in $[0, 1]$ was proposed to tackle shortcomings of BLEU in distant language pairs, where changes in word ordering deteriorates BLEU.

\section{Discussions}
\label{section:summary}

% We find that SentencePiece performs considerably well for out-of vocabulary data like named entities, reducing to transliteration in most of those settings.

The results using our systems for {\sc wat-2018} are presented  in Table \ref{table:results} (see some additional results in Table~\ref{table:transformer-results}). The first part of the table consists of results on combinations of datasets and augmentations. All values are for models trained from scratch. In the second part, the current leader board is indicated for comparison. Note that entries in this part don't correspond to a single submission, but the values corresponding to the best in the respective metric.

Our submission based on the combination National News and IITB corpus tops human evaluation in Hindi to English, and ranks second in  English to Hindi. We demonstrate the possibility of distilling knowledge of online available sources into a usable translation model. We successfully use the \textsc{convs2s} architecture along with SentencePiece to obtain results comparable to the top submissions. Our experiments also indicates data augmentation using backtranslation positively works for the Hindi-English pair.

\section{Additional Transformer Experiments}
\label{section:additional-transformer-experiments}

In this section, we present a set of experiments and results post WAT-2018 involving the Transformer Architecture \cite{vaswani2017attention}. Two variants of the architecture - Transformer Base and Transformer Big outperformed then state of the art \textsc{convs2s} models in the WMT German-English and French-English translation tasks.

We used the Transformer-Base architecture in further experiments with the National News + IITB corpus where \textsc{convs2s} performed the best, with the rest of the pipeline being kept same as described before. We went with the default hyperparameters provided by \emph{fairseq} framework - which did not give us impressive results.

Following Popel and Bojar \shortcite{popel2018training}, we modified the hyperparameters for initial warm-up steps of 16000 without any learning rate decay, starting from a learning rate of 0.25, followed by an exponential decay of learning rate. We also had to enable delayed gradient updates \cite{ott2018scaling} to simulate a larger batch on smaller GPU before the model demonstrated any learning. During inference time, we averaged checkpoints of the model at different epochs once the loss on the development set had plateaued to obtain better results than a single checkpoint.

\begin{table}[htbp]
\centering
\begin{tabular}{|l|c|c|c|}
\hline
Architecture & BLEU & RIBES & AM-FM \\ \hline
\textsc{convs2s} & 19.69 & 0.758365	& 0.699810 \\
Transformer & 21.10 & 0.771549	& 0.712200 \\
~+Averaging & \textbf{21.57} & \textbf{0.773923} & \textbf{0.712110}	 \\ 
% T-Big (avg) & 21.42 & 0.768969 & 0.707020	 \\
\hline
\end{tabular}
\caption{Transformer-Base vs \textsc{convs2s} on National News + IITB corpus, for English to Hindi direction.}
\label{table:transformer-results}
\end{table}

In Table \ref{table:transformer-results}, we compare the performance of the transformer with that \textsc{convs2s}. Consistent with observations in languages like German-English and French-English, the transformer network produces better results than \textsc{convs2s} on all metrics. The averaged model performs the best in all metrics in English to Hindi translation task, at the time of writing this paper.

\section{Observations}
\label{section:observations}
We believe that {\sc nmt} is a promising approach for Indian language machine translation for obtaining reasonably accurate solutions. Our initial results reported here confirms this. In addition, we believe, the popular data augmentation methods are effective and feasible for many low-resource machine translation settings. We see the direct
utility of the advances in {\sc nmt} for many western language pairs on English-Hindi in terms of ideas and architectures. At the same time, we also believe, there is much more to 
do for making them effective on Indian languages.
%extending and enhancing these attempts for other Indian languages and also for sup

\section*{Acknowledgments}
We thank the organizers for systematically setting up this task, and for the very useful resources.
We also thank the larger language processing group at IIIT Hyderabad for the encouragement, support and insights.

% Do not number the acknowledgment section. Do not include this section when submitting your paper for review.

\bibliography{main}
\end{document}